\documentclass[runningheads]{llncs}

\usepackage{amsthm}
\usepackage{eccv}

\usepackage{eccvabbrv}

\usepackage{graphicx}
\usepackage{booktabs}

\usepackage[accsupp]{axessibility}  

\usepackage{multirow}
\newcommand{\argmin}{\text{\normalfont argmin}}
\theoremstyle{plain}
\theoremstyle{definition}
\theoremstyle{remark}
\usepackage{algorithm}
\usepackage{algorithmic}

\usepackage{subcaption}
\DeclareCaptionSubType * [alph]{table}
\usepackage{hyperref}

\usepackage{hyperref}

\usepackage{orcidlink}

\begin{document}

\title{Learning Representation for Multitask Learning through Self-Supervised Auxiliary Learning} 

\titlerunning{Multitask Learning through Self-Supervised Auxiliary Learning}

\author{Seokwon Shin \inst{1,2}\orcidlink{0000-0002-1267-7196} \and
Hyungrok Do\inst{3}\orcidlink{0000-0001-5317-6809} \and
Youngdoo Son\inst{1,2,*}\orcidlink{0000-0002-1912-5853}}

\authorrunning{S. Shin et al.}

\institute{Department of Industrial and Systems Engineering, Dongguk University-Seoul, 30, Pildong-ro 1-gil, Jung-gu, Seoul, 04620, Republic of Korea \\
 \email{shinseokwon@dgu.ac.kr, youngdoo@dongguk.edu } \and
 Data Science Laboratory (DSLAB), Dongguk University-Seoul, 30, Pildong-ro 1-gil, Jung-gu, Seoul, 04620, Republic of Korea \and
Department of Population Health, NYU School of Medicine, New York, NY, 10016, United States \\ \email{doh03@nyu.edu} \\
* corresponding author} 


\maketitle
\begin{abstract}
Multi-task learning is a popular machine learning approach that enables simultaneous learning of multiple
related tasks, improving algorithmic efficiency and effectiveness. In the hard parameter sharing approach, an encoder shared through multiple tasks generates data representations passed to task-specific predictors. Therefore, it is crucial to have a shared encoder that provides decent representations for every and each task. However, despite recent advances in multi-task learning, the question of how to improve the quality of representations generated by the shared encoder remains open. To address this gap, we propose a novel approach called Dummy Gradient norm Regularization (DGR) that aims to improve the universality of the representations generated by the shared encoder. Specifically, the method decreases the norm of the gradient of the loss function with respect to dummy task-specific predictors to improve the universality of the shared encoder’s representations.
Through experiments on multiple multi-task learning benchmark datasets, we demonstrate that DGR effectively improves the quality of the shared representations, leading to better multi-task prediction performances. Applied to various classifiers, the shared representations generated by DGR also show superior performance compared to existing multi-task learning methods. Moreover, our approach takes advantage of computational efficiency due to its simplicity. The simplicity also allows us to seamlessly integrate DGR with the existing multi-task learning algorithms.
  \keywords{Multi-task learning \and Universality \and Regularization}
\end{abstract}

\section{Introduction}
\label{sec:intro}

Multi-task learning (MTL) is a machine learning approach that involves training a single model to handle multiple tasks simultaneously by sharing model parameters across the tasks, leading to better efficiency and a less complex model compared to having completely separate models for each task. MTL can potentially improve the quality of learned representations, which can benefit individual tasks. MTL is particularly useful in real-world applications where multiple tasks need to be performed simultaneously with limited resources. Representative examples include autonomous driving perception\cite{chowdhuri2019multinet,chen2018multi,ishihara2021multi}, defect detection\cite{sampath2023attention,li2021end,dong2021defect,shao2022pixel}, and pre-training techniques\cite{chen2021lottery,anderson2022improving,zhang2022task,bachmann2022multimae}. 

However, learning multiple tasks simultaneously can be a challenging problem, and optimizing the average loss over all tasks in MTL may not always result in satisfactory generalization performance \cite{caruana1998multitask,zhang2018overview}. Moreover, sharing the representation in MTL can lead to a challenge where some tasks are learned well, while others are overlooked due to differences in the loss scales and gradient magnitudes of various tasks, as well as the interference among them. As a result, some tasks may dominate the training process, leading to poorer performance on other tasks \cite{evgeniou2004regularized,ruder2017overview}. 

Various methods have been proposed to address these challenges and improve the generalization performance of MTL models. These methods include manipulating gradients to prevent their conflicts \cite{yu2020gradient,liu2021conflict,liu2021towards,lee2022multitask,desideri2012multiple,chen2018gradnorm,chen2020just}, properly weighting loss functions \cite{kendall2018multi,guo2018dynamic,liu2022auto,liu2019end,liu2019loss,liu2021towards}, and finding Pareto optimal solutions for MTL as a multi-objective optimization problem \cite{navon2022multi,lin2019pareto,desideri2012multiple,sener2018multi}. These approaches have contributed to our understanding of the challenges in MTL and provided valuable insight to improve its performance.

Despite all these recent advances in MTL, the question of how to improve the universality of the representations generated by the shared encoder has not been explored. In this paper, we present a novel method for improving the universality of the shared encoder as a form of regularization. To begin, we define \emph{universality} as the inverse of the difference between the loss of an optimal task-specific predictor and that of an untrained task-specific predictor, which is arbitrarily chosen. The underlying idea behind this definition is that the \emph{universal representations} generated by the encoder would allow any task predictor, whether optimal or arbitrary, to perform with equal efficacy. Moreover, we show that the universality is inversely proportional to the Frobenius norm of the gradient of the loss function with respect to the arbitrarily chosen predictors. This allows us to increase the universality just by adding a simple dummy gradient norm to the learning objective function. Owing to its simplicity, we integrate our approach with the existing MTL approaches and demonstrate that our approach boosts the baseline performances in most combinations.

In summary, the contributions of our work are:
\begin{itemize}
\renewcommand{\labelitemi}{$\bullet$}
\item defining the universality of the shared encoder for hard parameter sharing MTL framework as the inverse of the difference between the loss value of an optimal task-specific predictor and that of an arbitrary predictor;
\item showing that the universality is inversely proportional to the Frobenius norm of the gradient of the loss function with respect to an arbitrary predictor;
\item proposing Dummy Gradient norm Regularization (DGR), a novel regularization method to improve the universality of the shared encoder;
\item demonstrating that the DGR improves the performances of the existing MTL methods, as easily being integrated with them, through a series of experiments on various MTL benchmark datasets;
\item demonstrating that the DGR results in decent-quality representations of input data.
\end{itemize}

\section{Related Work}
The main challenge in hard parameter sharing MTL is \emph{negative transfer}, which refers to a situation in which the performance of a model on one task is adversely affected by another task. This can happen when the knowledge or representations learned on one task are not compatible with that of another task, leading to a conflict or mismatch. The previous studies tackled the negative transfer problem in several different ways: (1) loss weighting, (2) gradient manipulation, and (3) multi-objective optimization.

Loss weighting assigns different weights on loss functions for each task to minimize the negative transfer caused by the large variance in loss or gradient magnitudes per task \cite{kendall2018multi,guo2018dynamic,liu2022auto,liu2019end,liu2019loss,liu2021towards}. Methods in this category differ in how to determine the weight for each task. For example, \cite{kendall2018multi} determined the weights based on the uncertainty for each task, and \cite{liu2019end} used previous loss improvements to adjust the weights during training. Recently, \cite{liu2022auto} determined weights using a meta-learning scheme similar to MAML \cite{finn2017model} and showed excellent performance improvement.

Moreover, there exists a line of work that formulates MTL as a multi-objective optimization problem \cite{desideri2012multiple, sener2018multi, lin2019pareto, navon2022multi}. It is also known that the multi-objective optimization-based approaches mitigate the negative transfer problem. In addition, there exists an integrated approach that involves both loss weighting and gradient manipulation \cite{liu2021towards}.

Meanwhile, previous studies \cite{bengio2013representation,kolesnikov2019revisiting,kolesnikov2020big,rebuffi2017icarl} argued that the success of deep neural networks for computer vision is related to the quality of representation. 
In the case of MTL, how the shared representations are suitable for all tasks is one of the key components for achieving good performance across all tasks.
However, to the best of our knowledge, the universality of representations generated by a shared encoder in MTL has yet to be explored; even a concept of universality for MTL problem has never been considered. 

\begin{figure*}[!ht]
  \centering
  \includegraphics[width=1.0\columnwidth]{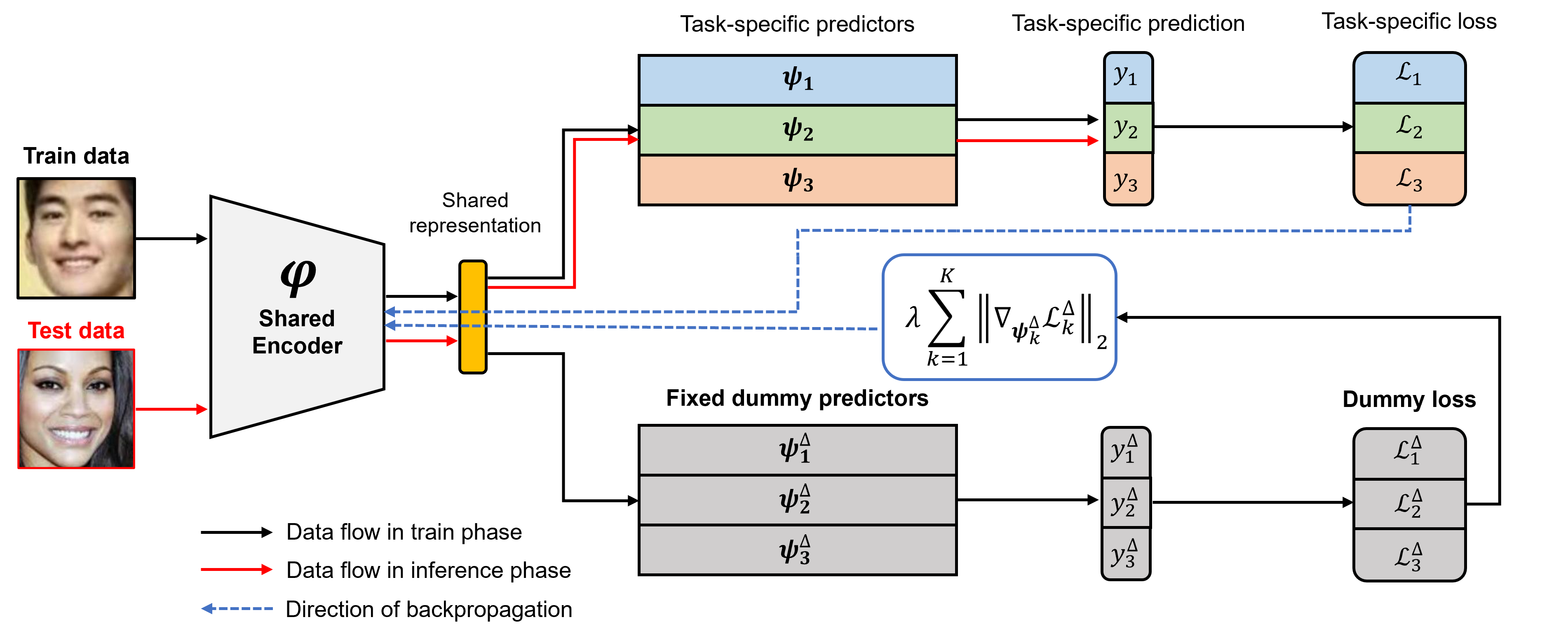}
  \caption{A schematic overview of the proposed DGR, which consists of a shared encoder, task-specific predictors, and task-specific dummy predictors. During the forward pass, task-specific predictors produce the actual prediction for each task, while the backward pass minimizes the sum of task-specific losses and encourages the universality of the shared encoder using dummy predictors. The black and red solid lines represent the forward pass during the training and inference phases, respectively, while the blue dashed line represents the direction of backpropagation for training.}
  \label{fig:OVERVIEW}
\end{figure*}

\section{Proposed Method}
This section introduces the dummy gradient norm regularization to improve the universality of shared representation in hard parameter sharing MTL setting. First, we introduce general problem setting for MTL. Then, we describe how DGR can improve the universality of the shared encoder. Figure \ref{fig:OVERVIEW} gives a schematic overview of the proposed method. 
\subsection{MTL Problem Definition}
Suppose we are given $K$ different prediction tasks $\mathcal{T}$ $=$ $\{\mathcal{T}_{1},\cdots, \mathcal{T}_{K}\}$ with training dataset $\mathcal{D}$ $=$ $\{(\mathbf{x}_{i}, \mathbf{y}_{1i},\cdots,\mathbf{y}_{Ki}) \in \mathcal{X} \times \mathcal{Y}_{1} \times \cdots \times \mathcal{Y}_{K}:i=1,\cdots,n\}$, where $\mathbf{x}_{i}$ is the input vector and $\mathbf{y}_{ki}$ is the true label for prediction task $\mathcal{T}_{k}$, for $k=1,\cdots,K$. The goal of MTL is to learn a model that achieves high average performance on all $K$ prediction tasks \cite{yu2020gradient}. We consider the hard parameter sharing MTL setting \cite{yu2020gradient,liu2021conflict,liu2022auto,liu2021towards,navon2022multi} where the prediction model $f$ consists of a shared encoder $\varphi$ and $K$ task-specific predictors $\Psi = \{\psi_{k}:k=1,\cdots,K\}$ so that $\psi_{k} ~\circ ~\varphi: \mathcal{X} \to \mathcal{Y}_{k}$ makes the predictions for task $\mathcal{T}_{k}$. We consider $\varphi$ and $\psi_{k}$ to be parameterized by $\theta_{\mathcal{E}}$ and $\theta_{\mathcal{T}_{k}}$, respectively, for each $k$, and let $\Theta = \{\theta_{\mathcal{E}}, \theta_{\mathcal{T}_{1}},\cdots, \theta_{\mathcal{T}_{K}}\}$ represent the set of all model parameters. Therefore, the MTL problem we consider is formulated as follows:
\begin{equation}\label{eq:1}
    \min_{\Theta} \sum_{i=1}^{n} \sum_{k=1}^{K}\mathcal{L}_{k}\big( \mathbf{y}_{ki}, ( \psi_{k} \circ \varphi)(\mathbf{x}_{i}) \big),
\end{equation}
where $\mathcal{L}_{k}: \mathcal{Y}_{k} \times \mathcal{Y}_{k} \to \mathbb{R}_{+}$ is a loss function for $\mathcal{T}_{k}$.

\subsection{Universality of Shared Encoder}\label{sec:unviversality}
There have been some attempts to measure the quality of representation in various fields \cite{bengio2013representation, hjelm2018learning,zhang2021understanding,wu2018unsupervised,tzeng2017adversarial,wang2019transferable}. 
They have investigated reconstruction errors, evaluated performance in downstream tasks, or evaluated transferability to new tasks.
However, these approaches require the intervention of downstream task predictors, which can potentially be computationally demanding and time consuming. Furthermore, such methods do not provide quality scores that can be directly incorporated into the learning objective function. Therefore, we propose a way to quantify the universality of the shared encoder. We define \emph{universality} as the inverse of the difference between the loss value of an optimal task-specific predictor and that of an arbitrarily task-specific predictor.

We first define optimal task-specific predictors.
\begin{definition}[Optimal Task-Specific Predictors]
    Given a shared encoder parameterized by $\theta_{\mathcal{E}}$, we denote the optimal task-specific predictor $\psi_{\theta^{*}_{\mathcal{T}_{k}}}$ for task $\mathcal{T}_{k}$ where
    \begin{equation}\label{eqn:optimal-predictor}
        \theta_{\mathcal{T}_{k}}^{*}|\theta_{\mathcal{E}} = \underset{\theta_{\mathcal{T}_{k}}}{\argmin} \sum_{i=1}^{n} \mathcal{L}_{k}(\mathbf{y}_{ki}, \big( \psi_{\theta_{\mathcal{T}_{k}}} \circ \varphi_{{\theta}_{\mathcal{E}}})(\mathbf{x}_{i}) \big).
    \end{equation}
    Note that $\theta_{\mathcal{T}_{k}}^{*}$ depends on $\theta_{\mathcal{E}}$, which implies that the optimal task-specific predictor varies as $\theta_{\mathcal{E}}$ perturbs.
\end{definition}

Apparently, the optimal task-specific predictors depend on the shared encoder, meaning that the difficulty of Problem \eqref{eqn:optimal-predictor} varies as $\theta_{\mathcal{E}}$ changes. That is, if the shared encoder generates poor representation, it might be hard to learn a good task-specific predictor. Moreover, if the shared encoder is biased towards certain (often simpler or easier) tasks, it may be difficult to learn the other tasks effectively. Thus, the representations generated by the shared encoder need to be sufficiently informative to help each predictor learn the task-specific features necessary for good performance. 

Therefore, we want the shared encoder to provide sufficiently good representations for all prediction tasks. We quantify the \emph{universality} of the shared encoder based on an arbitrarily chosen task-specific predictor, $\psi_{\theta_{\mathcal{T}_{k}}^{\Delta}}$. 

\begin{definition}[Universality of Shared Encoder]
    Given an arbitrary task-specific predictor $\psi_{\theta_{\mathcal{T}_{k}}^{\Delta}}$, we define the quality of representations provided by the shared encoder as follows:
    \begin{align}\label{eqn:QoE}
    \footnotesize
        \mathcal{U}(\theta_{\mathcal{E}} | \theta_{\mathcal{T}_{k}}^{\Delta} ) =  \Big[\sum_{i=1}^{n}\min_{\boldsymbol{\sigma}_k}\mathcal{L}_{k}\big(\sigma_{ki}y_{ki}, (\psi_{\theta_{\mathcal{T}_{k}}^{\Delta}} \circ \varphi_{\theta_{\mathcal{E}}})(\mathbf{x}_{i}) \big) \sum_{i=1}^{n}\mathcal{L}_{k}\big(y_{ki}, (\psi_{\theta_{\mathcal{T}_{k}}^{*} \circ}\varphi_{\theta_{\mathcal{E}}} )(\mathbf{x}_{i}) \big) \Big]^{-1}
    \end{align}
    where $\boldsymbol{\sigma}_{k}$ is a class permutation for the task $k$.
\end{definition}
The intuition behind this definition is that if the shared encoder produces universal representations of the input data, then an arbitrarily chosen non-trained predictor can achieve comparable performance to the optimal predictor. In other words, the universal representations should be remarkably informative and expressive, so that they can compensate for the lack of task-specific tuning of an arbitrarily initialized non-trained predictor. If the learned representation is universal for all tasks, it could greatly help each task-specific predictor achieve decent performance for its prediction task.

Our definition for universality shares a similarity with Sharpness-Aware Minimization (SAM)\cite{foret2020sharpness} in that they both minimize the discrepancy between the losses incurred by the parameters and alternative ones. However, SAM aims to explore flat minima by minimizing the difference between the losses with the given parameters and their vicinity, while our work finds universal representations by minimizing the difference between the losses with the optimal parameters and arbitrary ones. 

Although \eqref{eqn:QoE} is a reasonable quantification of the universality of the shared encoder, including $\mathcal{U}$ directly into the learning objective is computationally infeasible as it requires knowing the optimal task-specific predictors $\psi_{\theta_{\mathcal{T}_{k}}^{*}}$. The following theorem, which assumes the convexity of the loss function, allows us to have an indirect but feasible way of improving the universality of the shared encoder. The assumption of convexity has also been utilized in other multitask learning studies \cite{yu2020gradient,liu2021conflict} and has led to successful performance improvements.

\begin{theorem}[Inverse Proportionality]\label{thm:InvProp}
    Given a task-specific predictor $\psi_{\theta_{\mathcal{T}_{k}}^{\Delta}}$, if $\mathcal{L}_{k}$ is convex with respect to $\theta_{\mathcal{T}_{k}}^{\Delta}$, the universality of the shared encoder is inversely proportional to the Frobenius norm of the gradient of the loss function with respect to $\theta_{\mathcal{T}_{k}}^{\Delta}$, that is,
    \begin{equation}\label{eqn:InvProp}
        \mathcal{U}(\theta_{\mathcal{E}} | \theta_{\mathcal{T}_{k}}^{\Delta} ) \propto\left\| \nabla_{\theta_{\mathcal{T}_{k}}^{\Delta}} \sum_{i=1}^{n}\mathcal{L}_{k}\big(\mathbf{y}_{ki}, ( \psi_{\theta_{\mathcal{T}_{k}}^{\Delta}} \circ \varphi_{\theta_{\mathcal{E}}})(\mathbf{x}_{i}) \big) \right\|_{F}^{-1}.
    \end{equation}
\end{theorem}

Theorem \ref{thm:InvProp} allows us to improve the universality of the shared encoder by minimizing the Frobenius norm of the gradient of an arbitrary task-specific predictor. Therefore, we formulate our MTL problem as follows:
\begin{align} \label{eqn:learning_objective}
    \footnotesize
    \min_{\Theta} \sum_{k=1}^{K} \left[ \sum_{i=1}^{n}\mathcal{L}_{k}\big(\mathbf{y}_{ki}, (\psi_{\theta_{\mathcal{T}_{k}}}\circ \varphi_{\theta_{\mathcal{E}}} )(\mathbf{x}_{i}) \big) + \lambda \left\| \nabla_{\theta_{\mathcal{T}_{k}}^{\Delta}} \sum_{i=1}^{n}\mathcal{L}_{k}\big(\mathbf{y}_{ki}, (\psi_{\theta_{\mathcal{T}_{k}}^{\Delta}} \circ \varphi_{\theta_{\mathcal{E}}})(\mathbf{x}_{i}) \big)\right\|_{F} \right],
\end{align}

where $\lambda > 0$ is the hyperparameter that controls the tradeoff between the loss and the penalty term. Problem \eqref{eqn:learning_objective} can be optimized with typical minibatch stochastic gradient method provided that $\mathcal{L}_{k}$ is smoothly differentiable. The optimization steps to solve Problem \eqref{eqn:learning_objective} is summarized in Algorithm \ref{alg:1}.

Algorithm 1 involves the computation of a Hessian matrix of the loss function with respect to the parameters of the dummy predictor, which potentially requires substantial time and space complexity, depending on the number of parameters of the task-specific predictor. Therefore, for computational and memory efficiency, we adopt the finite difference approximation, which is widely used to approximate Hessian matrices  \cite{finn2017model,liu2018darts}. The details of the finite difference are given in Appendix.

\begin{algorithm}[t]
\caption{DGR for MTL}\label{alg:1}
\textbf{Input}: shared encoder $\varphi_{\theta_{\mathcal{E}}}$ and task-specific predictors $\{\psi_{\theta_{\mathcal{T}_{k}}}\}$, task-specific dummy predictors $\{\psi_{\theta_{\mathcal{T}_{k}}^{\Delta}}\}$, dataset $\mathcal{D}$, and hyperparamter $\lambda$\\
\textbf{Output}: Multitask prediction model $(\varphi_{\theta_{\mathcal{E}}}$, $\psi_{\theta_{\mathcal{T}_{k}}}).$

\begin{algorithmic}[1]
    \STATE $t \gets 0$, $(\theta_{\mathcal{E}}^{(0)}, \theta_{\mathcal{T}}^{(0)}) \gets \texttt{Initialize}(\theta_{\mathcal{E}}, \theta_{\mathcal{T}})$.
    \WHILE{\texttt{Not Converged}}
        \STATE $\mathcal{B} = \{(\mathbf{x}_{i},\mathbf{y}_{ki})\} \gets \texttt{MiniBatchSampler}(\mathcal{D},b)$.
        \STATE Feedforward encoder: $\mathbf{z}_{i} = \varphi_{\theta_{\mathcal{E}}^{(t)}}
        (\mathbf{x}_{i})$.
        \FOR{$k=1$ to $K$}
             \STATE Feedforward predictor: $\hat{\mathbf{y}}_{ki} = \psi_{\theta_{\mathcal{T}_{k}}^{(t)}}(\mathbf{z}_{i})$.
             \STATE Feedforward dummy predictor: $\hat{\mathbf{y}}_{ki}^{\Delta} = \psi_{\theta_{\mathcal{T}_{k}}^{\Delta}}(\mathbf{z}_{i})$.
            \STATE Compute objective value for the minibatch $\mathcal{B}$: $$ \hspace{-1.5em}\ell_{k}(\mathcal{B}) = \sum_{i=1}^{b}\Big[\mathcal{L}_{k}(\mathbf{y}_{ki}, \hat{\mathbf{y}}_{ki} ) + \lambda \|\nabla_{\small \theta_{\mathcal{T}_{k}}^{\Delta}}\mathcal{L}_{k}(\mathbf{y}_{ki}, \hat{\mathbf{y}}_{ki}^{\Delta})\|_{F}\Big].$$ 
            \STATE Update $\theta_{\mathcal{T}_{k}}$ by stochastic gradient descent:  $ \theta_{\mathcal{T}_{k}}^{(t+1)} \gets \theta_{\mathcal{T}_{k}}^{(t)} - \eta \nabla_{\theta_{\mathcal{T}_{k}}} \ell_{k}(\mathcal{B}). $
        \ENDFOR
        \STATE Update $\theta_{\mathcal{E}}$ by stochastic gradient descent:  $ \theta^{(t+1)}_{\mathcal{E}} \gets \theta^{(t)}_{\mathcal{E}} - \eta \nabla_{\theta_{\mathcal{E}}} \sum_{k=1}^{K} \ell_{k}(\mathcal{B}).$
        \STATE $t \gets t+1$.
    \ENDWHILE
    \STATE $(\theta_{\mathcal{E}}, \theta_{\mathcal{T}}) \gets (\theta_{\mathcal{E}}^{(t)}, \theta_{\mathcal{T}}^{(t)})$.

\end{algorithmic}
\end{algorithm}

\subsection{Choosing Task-Specific Dummy Predictor}\label{sec:choosing}
 As given in \eqref{eqn:InvProp}, $\mathcal{U}$ can be improved by decreasing the norm of the gradient of the loss function $\mathcal{L}_{k}$ given an arbitrary predictor $\psi_{\theta_{\mathcal{T}_{k}}^{\Delta}}$. In our implementation, we added a dummy predictor $\psi_{\theta_{\mathcal{T}_{k}}^{\Delta}}$ that has exactly the same architecture as the task-specific predictor $\psi_{\theta_{\mathcal{T}_{k}}}$ that is indeed used to make predictions. Then we initialize the dummy parameters $\theta_{\mathcal{T}_{k}}^{\Delta}$ randomly. The choice is arbitrary, but empirically we have confirmed that it helps improve performance. 

We fixed randomly initialized parameters for the dummy predictor during training for learning stability. However, there is a risk that the encoder may learn to adapt specifically to that fixed predictor. To prevent such adaptation, we utilized multiple dummy predictors instead of a single one. 
In our implementation, we employed three distinct, fixed dummy predictors.
In terms of employing additional decoders, our implementation draws parallels with Pseudo-Task Augmentation (PTA)\cite{meyerson2018pseudo}, which also employed multiple decoders across various tasks. However, while PTA aimed to train a shared encoder to solve the same tasks in multiple ways, our focus lies in enhancing the shared encoder's universality through gradient regularization with multiple fixed decoders.

Additionally, in terms of utilizing auxiliary tasks, our implementation is similar to previous studies on auxiliary learning \cite{navon2020auxiliary, du2018adapting, liu2019self, shamsian2023auxiliary}, which aimed to learn useful representations for specific primary tasks. However, while these studies have centered on determining or creating beneficial auxiliary tasks, to be learned alongside primary tasks, DGR focuses on regularizing the gradient norm obtained by dummy encoders with arbitrarily chosen and fixed parameters.

\section{Experiments}

We performed a series of experiments to demonstrate the utility and efficiency of the proposed DGR method on several multi-task benchmark datasets. We first apply our method to the image classification task of \textsc{UTKFace} \cite{zhifei2017cvpr} compiled as a multi-task learning problem. Then, we further investigate the quality of representations produced by the shared encoder learned with our approach. Subsequently, we tackle multi-task scene understanding problems on \textsc{NYUv2} \cite{silberman2012indoor} and \textsc{CityScapes} \cite{cordts2016cityscapes} datasets.

\subsection{Competitive Methods}

In our experiments, we compared our method with two baseline methods, single-task models ($\textsc{Single}$) and a vanilla multi-task model ($\textsc{Vanilla}$), and existing MTL methods. We considered loss weighting approaches that have demonstrated decent performance in previous studies, such as \textsc{Uncertainty} weighting \cite{kendall2018multi}, Dynamic Weight Averaging (\textsc{DWA}) \cite{liu2019end}, and \textsc{Auto-$\lambda$} \cite{liu2022auto}. Furthermore, we compared the proposed method with four other methods, including gradient conflict and multi-objective optimization: \textsc{PCGrad} \cite{yu2020gradient}, \textsc{CAGrad} \cite{liu2021conflict}, \textsc{IMTL} \cite{liu2021towards}, and \textsc{Nash-MTL} \cite{navon2022multi}.
We further integrated our method with each competitive MTL method (presented as +Ours) to investigate how our DGR can improve performance from the existing approaches.

\subsection{Multi-task Image Classification on \textsc{UTKFace}}
We evaluated DGR on the \textsc{UTKFace} dataset with three image classification tasks: age prediction, gender classification, and ethnicity classification. To facilitate a more straightforward interpretation of the experiment results, we transformed it into a classification task by discretizing ages into seven classes: 0-9, 10-19, 20-29, 30-39, 40-49, 50-59, and 60+.  

\noindent \textbf{Experiment Settings.} In our experiments on the \textsc{UTKFace} dataset, we used the standard hard parameter sharing structure known as \textsc{Split}, the same as in \cite{liu2022auto}. The \textsc{Split} structure is a general structure for hard parameter sharing settings, where each task-specific predictor makes predictions using only representations obtained through the shared encoder's parameters.

For all methods, we used ResNet50 \cite{he2016deep} as the backbone architecture for the shared encoder and two fully connected layers as each task-specific predictor. For optimization, we used Adam\cite{kingma2014adam} optimizer for all methods and performed a grid search on $\lambda \in {\{ 10^{-5},10^{-6},10^{-7} \}}$ to determine the optimal values of the hyperparameter for the proposed method.

\noindent \textbf{Evaluation Metrics.}
We reported the test accuracy on every classification task for the \textsc{UTKFace} dataset. All performances were averaged over three independent trials. Furthermore, we reported the relative improvement in performance $\Delta_{\text{MTL}}$ compared to the single task model in the following experiments \cite{maninis2019attentive}. $\Delta_{\text{MTL}}$ is defined as follow:
\begin{equation}\label{eq:12}
    \Delta_{\text{MTL}} = \frac{1}{K}\sum^{K}_{k=1}(-1)^{\varsigma_{k}}(M_{m,k}-M_{b,k})/M_{b,k}
\end{equation}
where $\varsigma_{k}$ is the indicator that takes 1 if the performance metric $M_{k}$ for $\mathcal{T}_{k}$ is the lower the better 0 otherwise. $m$ and $b$ represent multi-task model and single-task baseline model, respectively.

\begin{table}[!t]
    \caption{Test accuracies averaged over three random trials on the \textsc{UTKFace} dataset. Boldface and underline denote the best and second-best, respectively.}
    \setlength{\tabcolsep}{3pt}
    
      \label{tab:UTKFace}
      \centering
      \resizebox{5.5cm}{!}
  {
    \begin{tabular}{c|ccc|c}
    \toprule[1.5pt]
    Method      & Age & Gender & Ethnicity   &$\Delta_{\text{MTL}}$ $\uparrow$  \\ \midrule[1.5pt]
    \textsc{Single}     & 0.5233 & 0.8964 & 0.7728 & - \\\midrule
    \textsc{Vanilla}    & 0.5391 & 0.9153 & 0.7948 & +2.66\% \\
    + Ours               & \textbf{0.5605} & 0.9019 & 0.7942 & +3.50\% \\\midrule
    \textsc{DWA}        & 0.5370 & 0.9148 & 0.7989 & +2.68\% \\
    + Ours               & 0.5336 & 0.9135 & \underline{0.8056} & +2.71\% \\\midrule
    \textsc{Uncertainty} & 0.5385 & 0.9137 & 0.7961 & +2.62\% \\
    + Ours               & 0.5455 & 0.9118 & 0.7988 & +3.11\% \\\midrule
    \textsc{Auto}-$\lambda$ & 0.5405 & 0.9171 & 0.7965 & +2.89\% \\
    + Ours               & 0.5503 & 0.9163 & 0.8014 & +3.69\% \\\midrule
    \textsc{PCGrad}     & 0.5417 & 0.9134 & 0.7960 & +2.80\%\\
    + Ours               & 0.5431 & 0.8999 & 0.7986 & +2.50\% \\\midrule
    \textsc{CAGrad}     & 0.5390 & \textbf{0.9239} & 0.7994 & +3.17\% \\
    + Ours               & \underline{0.5510} & 0.9169 & \textbf{0.8094} & \textbf{+4.11\%} \\\midrule
    \textsc{IMTL}       & 0.5250 & \underline{0.9173} & 0.8039 & +2.23\% \\
    + Ours               & 0.5430 & 0.9020 & 0.8051 & +2.86\% \\\midrule
    \textsc{Nash-MTL}   & 0.5485 & 0.9172 & 0.8021 & +3.64\% \\
    + Ours               & 0.5545 & 0.9063 & 0.8055 & \underline{+3.77\%}  \\\bottomrule[1.5pt]
    \end{tabular}}
\end{table}

\noindent \textbf{Results.} Table \ref{tab:UTKFace} presents the image classification performance of the proposed method, baselines, and other MTL methods on \textsc{UTKFace} dataset. In addition, we also reported the performance of baselines and other MTL methods combined with our method. Across all tasks, the MTL methods, including DGR, outperformed the single-task model. Notably, both when independently employed and integrated with the other MTL methods, DGR improved performance in two of the three tasks—specifically, in the challenging age prediction and ethnicity classification tasks, which involve a larger number of classes than the gender classification task. Moreover, combined with DGR, $\Delta_{MTL}$ of almost all MTL methods, except \textsc{PCGrad}, increased.
This result emphasizes the efficacy of DGR in enhancing the universality of representations, thereby facilitating the obtaining of suitable representations for diverse and intricate tasks.

\begin{figure}[!t]
  \centering
  \includegraphics[width=0.6\columnwidth]{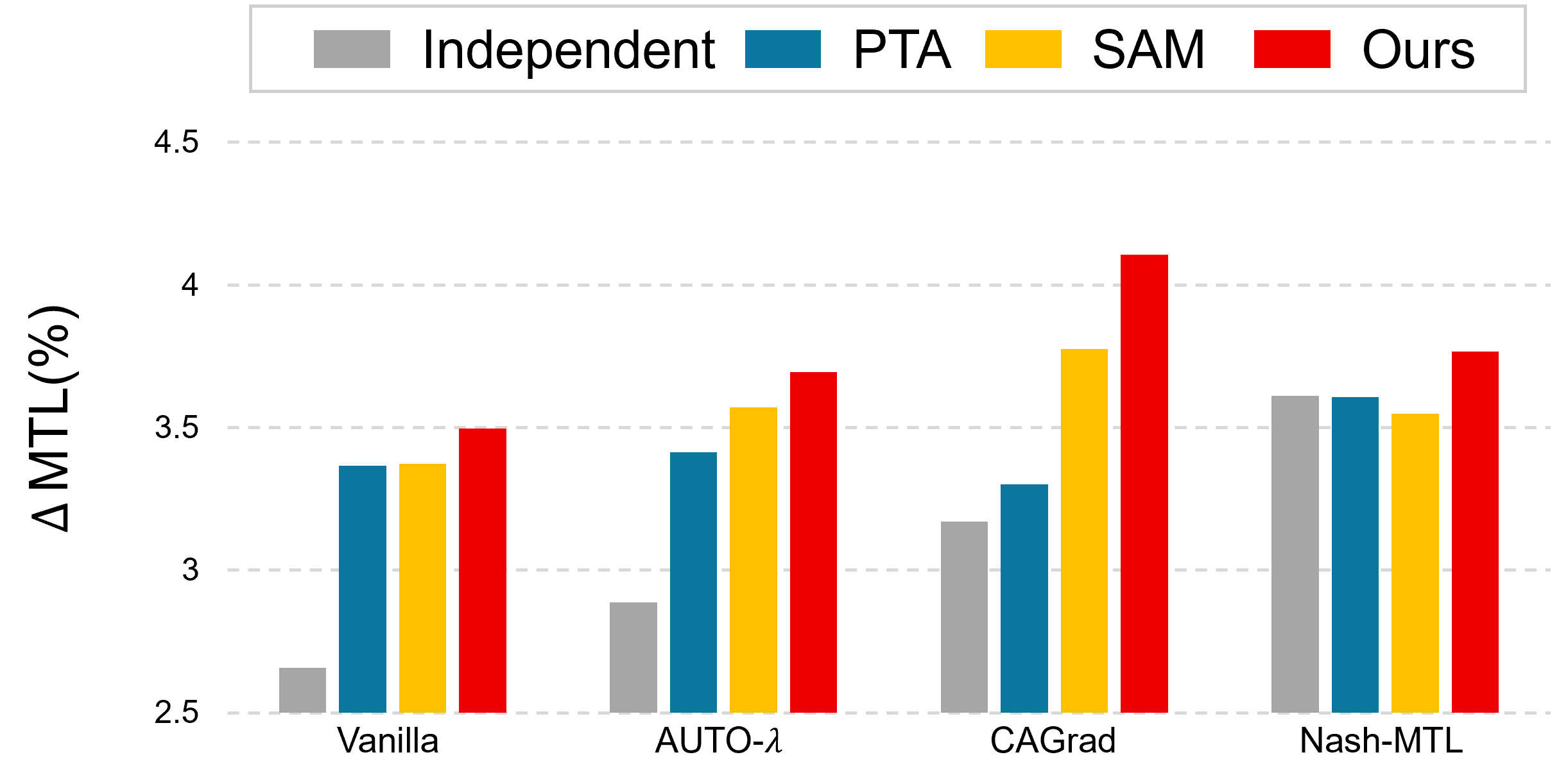}
  \caption{Comparison result of average $\Delta_{\text{MTL}}$ over three trials on the \textsc{UTKFace} dataset upon integrating each method into \textsc{SAM}, \textsc{PTA}, and Ours. The bars colored in gray indicate the results obtained using each method independently.}
  \label{fig:SAMPTA}
\end{figure}

We also conducted comparative experiments with DGR against \textsc{SAM}\cite{foret2020sharpness} and \textsc{PTA}\cite{meyerson2018pseudo}, which share some similarities with our work. We compared average $\Delta_{\text{MTL}}$ of each method when they are used both independently and combined with the three MTL methods that demonstrated the highest performances in Table \ref{tab:UTKFace}: \textsc{Auto}-$\lambda$, \textsc{CAGrad} and \textsc{Nash-MTL}. As shown in Figure \ref{fig:SAMPTA}, although DGR, SAM, and PTA showed consistent performance improvement when integrated into the MTL methods, the proposed DGR showed the best performance improvement.
The analysis on computational cost is also reported in Appendix.

\subsection{Quality of Shared Representation} \label{sec:qual}

According to our theoretical finding, a universal representation should enable any classifier to perform well. This includes relatively simpler classifiers built with the representations. We conducted an experiment to determine whether the shared representation learned through DGR, which was aimed at achieving a universal shared representation, exhibited greater universality compared to the shared representations learned through other methods.

In particular, we conducted a three-step experiment to assess the quality of shared representations on the dataset \textsc{UTKFace}. In the first step, we trained a model using each MTL method. Then, we froze the trained shared encoder and substituted the original classifier with a simpler alternative, such as decision tree (DT), support vector machine (SVM), and $k$-nearest neighbor (kNN). Finally, we evaluated the performance of these classifiers. This experiment is similar to previous evaluations of representation learning \cite{hjelm2018learning, bengio2013representation}, with the only variation being the algorithm used for the task-specific predictor.

\begin{figure}[t!]
\centering
\subfloat[]{\includegraphics[width=0.4\columnwidth]{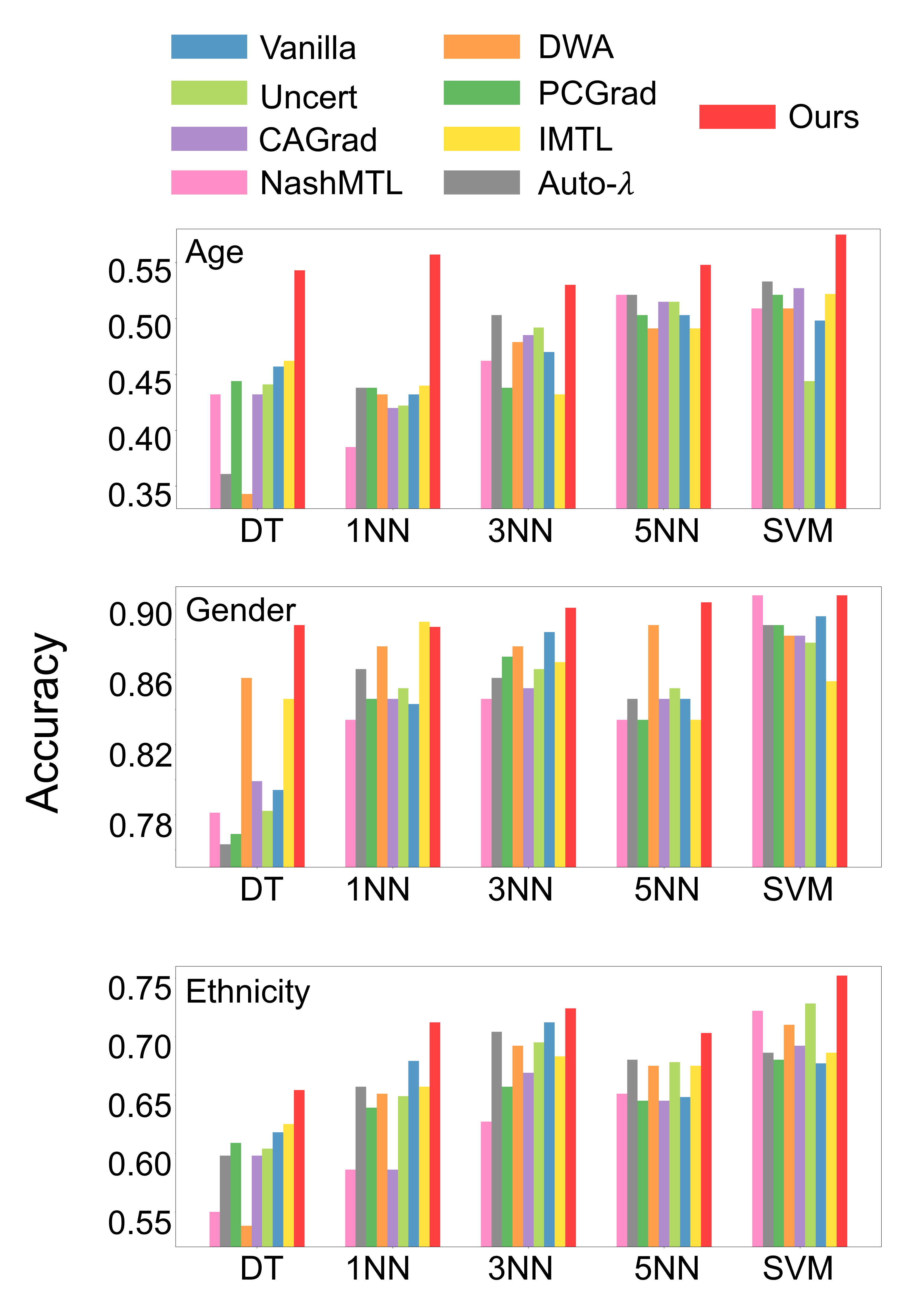}  \label{fig:simple}} 
\subfloat[]{\includegraphics[width=0.6\columnwidth]{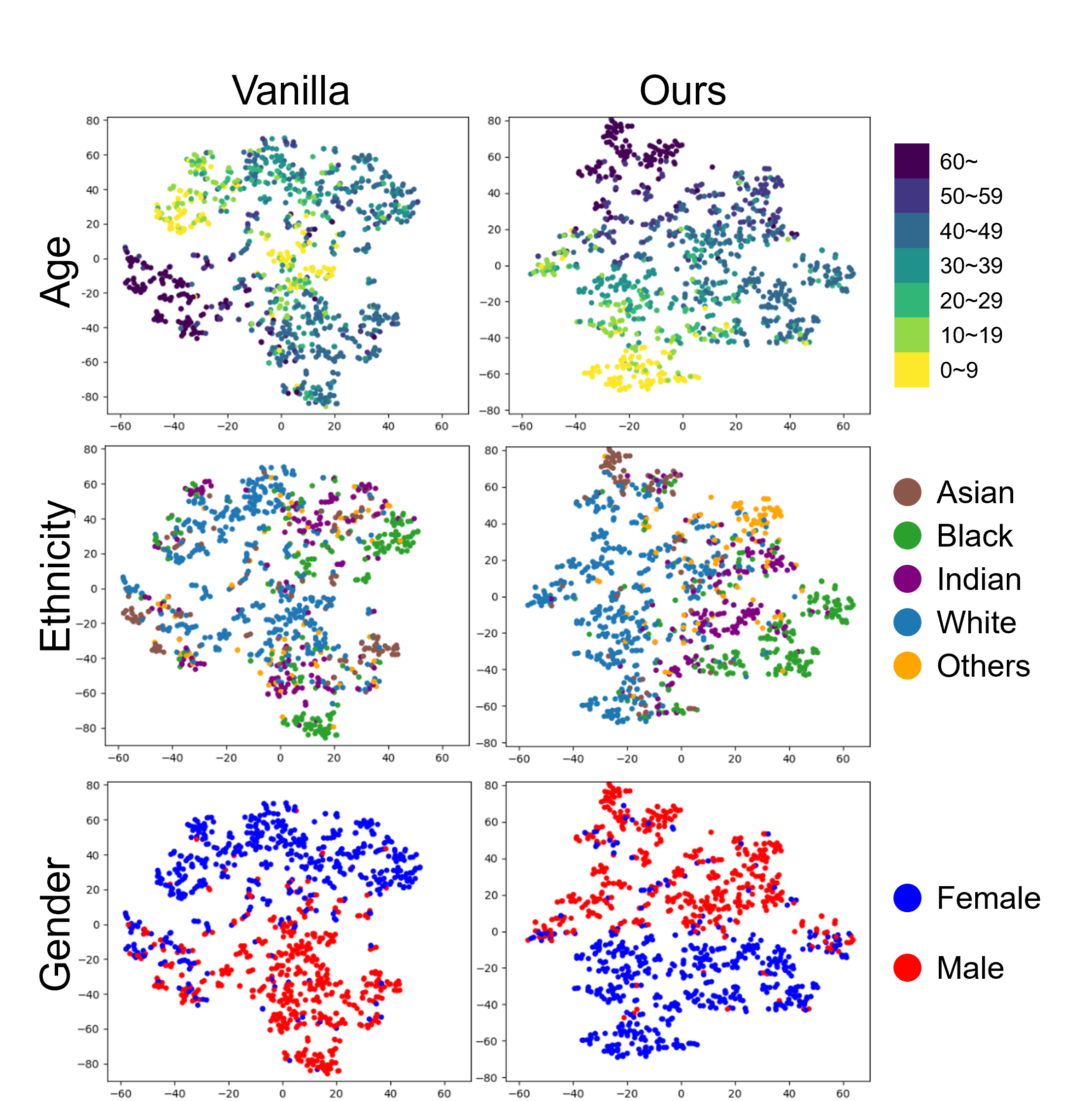}
\label{fig:t-SNE}}
\caption{(a) Average test performance over 3 trials on the \textsc{UTKFace} dataset, where the learned representation using MTL was fixed, and relatively simple classifiers were used to replace the original classifier. (b) The t-SNE visualization results of shared representations generated by the \textsc{Vanilla} (left) and DGR (right). Each row corresponds to a specific task.}
\label{some example}
\end{figure}

As shown in Figure \ref{fig:simple}, the shared representations learned through DGR consistently demonstrated superior performance with the simple task-specific classifier across all tasks. In particular, the performance advantage was more pronounced in the age prediction task, which has the largest number of classes. The proposed method was also less sensitive to the choice of classifiers.

Furthermore, we visualized the shared representations with t-SNE to verify the effect of DGR on their universality. Figure \ref{fig:t-SNE} shows the visualization results of the shared representation by the \textsc{Vanilla} and DGR. Similar to the experiment with simple classifiers, notable improvements were observed in both ethnicity and age prediction tasks. Specifically, in the ethnicity classification task, the representations generated by DGR were more clustered by their classes compared to those of \textsc{Vanilla}. Notably, for the age classification, the representations generated by DGR were not only clustered by classes but also aligned by age order, whereas those generated by the \textsc{Vanilla} were rather dispersed.

\subsection{Visual Scene Understanding on \textsc{NYUv2} and \textsc{CityScapes}}
We evaluated DGR on scene understanding tasks with the two standard datasets, \textsc{NYUv2} \cite{silberman2012indoor} and \textsc{CityScapes} \cite{cordts2016cityscapes}. In the experiments on the \textsc{NYUv2} dataset, we trained on three tasks: surface normal prediction, 13-class semantic segmentation, and depth prediction. In the experiments on the \textsc{CityScapes} dataset, we also trained on three tasks: 19-class semantic segmentation, 10-class part segmentation \cite{de2021part}, and disparity estimation, with the same setting as in \cite{kendall2018multi}. The experiments on these datasets are common for verification in MTL studies. 

\noindent \textbf{Experiment Settings.}
For the experiments on scene understanding tasks, we used the state-of-the-art multi-task architecture {MTAN}\cite{liu2019end} that has been primarily applied to the same experiments in previous studies, along with the standard structure {Split} used in image classification experiments. The {MTAN} structure differs from the {Split} by incorporating task-specific attention to obtain shared representations. We used ResNet-50 as the backbone architecture for all methods. Furthermore, we used the SGD optimizer for all methods except \textsc{Nash-MTL}, where we used Adam \cite{kingma2014adam} to achieve stable convergence of the algorithm \cite{navon2022multi}. We performed a grid search to determine the optimal values of $\lambda$, the hyperparameter of the proposed method, as the same as image classification experiments.

\begin{table*}[!t]
\caption{Average test performance over three trials on the \textsc{NYUv2} dataset with MTL methods in {Split} (left) and {MTAN} (right) multi-task architectures. The bold and underline denote the best and second best, respectively. The arrow next to metrics indicates whether the metric is higher or lower the better.}
\renewcommand{\arraystretch}{.9}
  \label{tab:NYU}
  \centering
  \resizebox{11cm}{!}
  {
\begin{tabular}{c|cccc|cccc}
\toprule[1.5pt]
\multirow{3}{*}{Method} & \multicolumn{4}{c|}{\textbf{Split}}                                                                                                    & \multicolumn{4}{c}{\textbf{MTAN}}                                                                                                      \\ \cmidrule(l){2-9} 
                        & \multicolumn{1}{c}{Normal} & \multicolumn{1}{c}{Semantic Seg.} & \multicolumn{1}{c|}{Depth} & \multirow{2}{*}{$\Delta_{\text{MTL}}$ $\uparrow$}        & \multicolumn{1}{c}{Normal} & \multicolumn{1}{c}{Semantic Seg.} & \multicolumn{1}{c|}{Depth} & \multirow{2}{*}{$\Delta_{\text{MTL}}$ $\uparrow$}     \\ 
                        & \multicolumn{1}{c}{(mDist.) $\downarrow$} & \multicolumn{1}{c}{(mIoU) $\uparrow$} & \multicolumn{1}{c|}{(aErr.) $\downarrow$} & & \multicolumn{1}{c}{(mDist.) $\downarrow$} & \multicolumn{1}{c}{(mIoU) $\uparrow$} & \multicolumn{1}{c|}{(aErr.) $\downarrow$} &       \\ \midrule[1.5pt]
\textsc{Single}        & 22.40 & 43.37 & \multicolumn{1}{c|}{52.24} & -        & 22.40 & 43.37 & \multicolumn{1}{c|}{52.24} & -        \\ \midrule
\textsc{Vanilla}       & 24.19 & 46.02 & \multicolumn{1}{c|}{41.36} & +6.32\%  & 24.05 & 46.05 & \multicolumn{1}{c|}{41.03} & +6.55\%  \\
+ Ours                 & 23.98 & 47.11 & \multicolumn{1}{c|}{40.11} & +8.26\%  & 24.04 & 46.24 & \multicolumn{1}{c|}{39.98} & +7.51\%  \\ \midrule
\textsc{DWA}           & 24.20 & 45.84 & \multicolumn{1}{c|}{41.58} & +6.02\%  & 24.06 & 46.47 & \multicolumn{1}{c|}{41.48} & +6.71\%  \\
+ Ours                 & 24.18 & 46.43 & \multicolumn{1}{c|}{41.08} & +6.82\%  & 24.10 & 46.81 & \multicolumn{1}{c|}{41.37} & +7.24\%  \\ \midrule
\textsc{Uncertainty}   & 24.24 & 46.46 & \multicolumn{1}{c|}{41.04} & +6.78\%  & 23.97 & 46.96 & \multicolumn{1}{c|}{41.02} & +7.57\%  \\
+ Ours                 & 24.34 & 46.52 & \multicolumn{1}{c|}{40.80} & +6.83\%  & 23.89 & 46.23 & \multicolumn{1}{c|}{40.00} & +7.28\%  \\ \midrule
\textsc{Auto-$\lambda$}& 23.60 & 48.20 & \multicolumn{1}{c|}{39.83} & +9.85\%  & 23.43 & 48.10 & \multicolumn{1}{c|}{39.21} & +10.02\% \\
+ Ours                 & 23.60 & 48.04 & \multicolumn{1}{c|}{39.45} & +9.96\% & 23.32 & 48.19 & \multicolumn{1}{c|}{39.03} & +10.50\% \\ \midrule
\textsc{PCGrad}        & 24.55 & 45.41 & \multicolumn{1}{c|}{42.32} & +4.70\%  & 24.07 & 46.71 & \multicolumn{1}{c|}{39.79} & +5.41\%  \\
+ Ours                 & 24.03 & 46.71 & \multicolumn{1}{c|}{41.20} & +7.19\%  & 23.95 & 46.86 & \multicolumn{1}{c|}{39.73} & +7.42\%  \\ \midrule
\textsc{CAGrad}        & 23.39 & 46.79 & \multicolumn{1}{c|}{40.55} & +8.61\%  & 23.23 & 47.48 & \multicolumn{1}{c|}{39.92} & +9.38\%  \\
+ Ours                 & 23.28 & 47.48 & \multicolumn{1}{c|}{39.81} & +9.78\%  & 22.98 & 48.49 & \multicolumn{1}{c|}{38.94} & +11.56\%  \\ \midrule
\textsc{IMTL}          & \underline{22.45} &\textbf{50.11} &\multicolumn{1}{c|}{\underline{38.69}}  & \underline{+13.75\%} &\underline{22.53} & \underline{49.84} &\multicolumn{1}{c|}{\underline{38.43}}& \underline{+13.43}\% \\ 
+ Ours                 &\textbf{22.37} & \underline{49.62} &\multicolumn{1}{c|}{\textbf{37.70}}& \textbf{+14.13}\% & \textbf{22.39} & \textbf{50.04} & \multicolumn{1}{c|}{\textbf{38.18}} & \textbf{+14.42}\%  \\ \midrule

\textsc{Nash-MTL}      & 23.60 & 48.20 & \multicolumn{1}{c|}{39.83} & +9.85\% & 23.43 & 48.10 & \multicolumn{1}{c|}{39.84} & +10.02\% \\ 
+ Ours                 & 22.54 & 48.25 & \multicolumn{1}{c|}{ 39.53} &+11.65\% & 22.66 & 48.38 & \multicolumn{1}{c|}{39.59} & +11.75\%  \\ \bottomrule[1.5pt] 
\end{tabular}
}
\end{table*}
\begin{table*}[hbt!]
\caption{Average test performance over three trials on the \textsc{CityScapes} dataset with MTL methods in {Split} (left) and {MTAN} (right) multi-task architectures. Bold and underline denote the best and second-best, respectively. The arrow next to metrics indicates whether the metric is higher or lower the better.}
\renewcommand{\arraystretch}{.9}
  \label{tab:CITY}
  \centering
  \resizebox{11cm}{!}
  {
\begin{tabular}{c|cccc|cccc}
\toprule[1.5pt]
\multirow{3}{*}{Method} & \multicolumn{4}{c|}{\textbf{Split}}                                                                                                    & \multicolumn{4}{c}{\textbf{MTAN}}                                                                                                      \\ \cmidrule(l){2-9} 
                        & \multicolumn{1}{c}{Semantic Seg.} & \multicolumn{1}{c}{Part Seg.} & \multicolumn{1}{c|}{Disparity} & \multirow{2}{*}{$\Delta_{\text{MTL}}$ $\uparrow$}        & \multicolumn{1}{c}{Semantic Seg.} & \multicolumn{1}{c}{Part Seg.} & \multicolumn{1}{c|}{Disparity} & \multirow{2}{*}{$\Delta_{\text{MTL}}$ $\uparrow$}     \\ 
                        & \multicolumn{1}{c}{(mIoU) $\uparrow$} & \multicolumn{1}{c}{(mIoU) $\uparrow$} & \multicolumn{1}{c|}{(aErr.) $\downarrow$} & & \multicolumn{1}{c}{(mIoU) $\uparrow$} & \multicolumn{1}{c}{(mIoU) $\uparrow$} & \multicolumn{1}{c|}{(aErr.) $\downarrow$} &       \\ \midrule[1.5pt]
\textsc{Single}   & 56.20 & 52.74 & \multicolumn{1}{c|}{0.84} & -        
                  & 56.20 & 52.74 & \multicolumn{1}{c|}{0.84} & -        \\ \midrule
\textsc{Vanilla}  & 56.29 & 52.31 & \multicolumn{1}{c|}{0.75} & +3.35\%  
                  & 56.64 & 52.52 & \multicolumn{1}{c|}{0.74} & +4.09\%  \\
+ Ours            & 56.27 & 52.45 & \multicolumn{1}{c|}{0.74} & +3.83\%  
                  & 57.13 & 51.88 & \multicolumn{1}{c|}{0.73} & +4.37\%  \\ \midrule
\textsc{DWA}      & 55.63 & 52.22 & \multicolumn{1}{c|}{0.72} & +4.10\%  
                  & 55.88 & 52.51 & \multicolumn{1}{c|}{0.72} & +4.43\%  \\
+ Ours            & 55.84 & 52.39 & \multicolumn{1}{c|}{0.72} & +4.33\%  
                  & 55.65 & 52.54 & \multicolumn{1}{c|}{0.72} & +4.31\%  \\ \midrule
\textsc{Uncertainty}   & 56.77  & 54.59 & \multicolumn{1}{c|}{0.77} & +4.29\%  
                       & 56.66  & 55.03 & \multicolumn{1}{c|}{0.76} & +4.89\%  \\
+ Ours                 & 56.76  & \underline{55.46} & \multicolumn{1}{c|}{0.78}  & +4.43\%  
                       & 56.59  & 55.10 & \multicolumn{1}{c|}{0.75} & +5.29\%  \\ \midrule
\textsc{Auto-$\lambda$} & 55.95 & 52.94 & \multicolumn{1}{c|}{\textbf{0.70}} & +5.53\%  
                       & 55.97  & 52.67 & \multicolumn{1}{c|}{\underline{0.69}} & +5.77\% \\
+ Ours                 & 57.22  & 53.55 & \multicolumn{1}{c|}{\underline{0.71}} & +6.28\% 
                       & 56.76  & 53.29 & \multicolumn{1}{c|}{\textbf{0.69}} & \underline{+6.63\%} \\ \midrule
\textsc{PCGrad}        & 55.40  & 52.53 & \multicolumn{1}{c|}{0.72} & +4.15\%  
                       & 55.53  & 52.83 & \multicolumn{1}{c|}{0.72} & +4.42\%  \\
+ Ours                 & 56.04  & 52.53 & \multicolumn{1}{c|}{0.72}   & +4.53\%  
                       & 56.61  & 52.75 & \multicolumn{1}{c|}{0.72}   & +5.01\%  \\ \midrule
\textsc{CAGrad}        & 56.55  & 55.05 & \multicolumn{1}{c|}{0.71} & \underline{+6.83}\%  
                       & 56.36  & 55.32 & \multicolumn{1}{c|}{0.72} & +6.49\%  \\
+ Ours                 & 57.05  & 55.16 & \multicolumn{1}{c|}{0.71} & \textbf{+7.19}\%  
                       & 56.66  & \textbf{56.07} & \multicolumn{1}{c|}{0.72} & \textbf{+7.14}\%  \\ \midrule
\textsc{IMTL}          & \underline{58.33} & 55.43 & \multicolumn{1}{c|}{0.76} & +6.14\% 
                        &\textbf{58.67} & \underline{55.77} & \multicolumn{1}{c|}{0.76}    & +6.55\% \\ 
+ Ours                  &  \textbf{58.46}  &  55.33 & \multicolumn{1}{c|}{ 0.75} &  +6.55\% 
                        & \underline{58.12}  & 55.73 & \multicolumn{1}{c|}{0.75} & +6.60\%  \\ \midrule
\textsc{Nash-MTL}       & 57.65  & \textbf{55.59} & \multicolumn{1}{c|}{0.75}  & +6.23\% 
                        & 57.64  & 55.09 & \multicolumn{1}{c|}{0.74}  & +6.31\% \\ 
+ Ours                  & 57.78 & 55.04 & \multicolumn{1}{c|}{ 0.74} &  +6.36\% 
                        & 57.77  & 55.60 & \multicolumn{1}{c|}{0.74}  & +6.71\%   \\ \bottomrule[1.5pt]
\end{tabular}}
\end{table*}

\noindent \textbf{Evaluation Metrics.} 
We followed the standard evaluation protocol that has been used in previous studies \cite{yu2020gradient,liu2021conflict,liu2022auto,liu2021towards,navon2022multi}. We evaluated normal, depth, and semantic segmentation tasks via absolute error (aErr.), mean angle distances (mDist.), and mean intersection over union (mIoU), respectively, for the experiments on the \textsc{NYUv2} dataset. Similar to this way, we evaluated two segmentation tasks, semantic and part, through mean intersection over union (mIoU) and the disparity estimation task via absolute error (aErr.) for the experiments on the \textsc{CityScapes} dataset.

\noindent \textbf{Results.}  
Tables \ref{tab:NYU} and \ref{tab:CITY} show the performance in the scene understanding experiments on the \textsc{NYUv2} and the \textsc{CityScapes} datasets, respectively.

On the \textsc{NYUv2} dataset, DGR showed similar or better performance than common benchmarks in recent MTL studies such as \textsc{DWA}, \textsc{Uncertainty}, and \textsc{PCGrad}. Notably, these results were achieved without loss balancing or gradient manipulation methods. Moreover, combining DGR with other methods resulted in performance improvements, and the best performance was achieved when \textsc{IMTL} was combined with the proposed method.

On the \textsc{CityScapes} dataset, DGR improved the performance of almost all MTL methods, except for \textsc{DWA}. Combining \textsc{CAGrad} with DGR led to the best performance.

These results confirmed that the DGR improves performance in the scene understanding tasks across most cases, even though the degrees of improvement vary by several factors, including datasets, model structures, tasks, and base methods.
\subsection{Sensitivity Analysis} 
\begin{table}[!htb] 
    \caption{\textbf{(a)} Experimental results with varying numbers of dummy decoders ($d$) and primary tasks (numbers in parentheses indicate the number of primary tasks). The final row shows the experimental result when DGR is utilized for single-task learning. For MTL, $\Delta_{\text{MTL}}$ is reported, while in single-task learning, top-1 accuracy is reported.  \textbf{(b)} Experimental results on the \textsc{NYUv2} and the \textsc{CityScapes} datasets when DGR utilized with the SWIN backbone. \textbf{(c)} Experimental results on on the \textsc{Pascal} dataset.}
    \begin{subtable}{.5\linewidth}
      \centering
      
      \resizebox{5.8cm}{!}{
        \begin{tabular}{c|c|ccc}
    \toprule[1.5pt]
        Dataset  & \textsc{Vanilla} & $d=1$ & $d=3$ & $d=5$     \\ \midrule[1.5pt]
    \textsc{NYUv2 (1)}    &  +0.00\% & +0.05\% & +0.07\% & -0.25\% \\\midrule
    \textsc{NYUv2 (2)}      &  +6.76\% & +6.84\% & +6.93\% & +6.73\%     \\\midrule
    \textsc{NYUv2 (3)}      &  +6.32\% & +7.13\% & +8.05\% & +7.96\%     \\\midrule[1.5pt]
    \textsc{Cityscapes (1)}    &  +0.00\% & -0.04\% & +0.02\% & +0.03\% \\\midrule
    \textsc{Cityscapes (2)}   & +3.24\% & +3.30\% & +3.33\% &+3.28\%\\\midrule
    \textsc{Cityscapes (3)}    &  +3.37\% & 3.72\% & +3.83\% & +3.66\% \\\midrule[1.5pt]
    \textsc{ImageNet}       &  76.13\% & 76.30\% & 76.50\% & 75.98\% \\\bottomrule[1.5pt]
    \end{tabular}}
    \caption{}\label{subtab:t1}
    \end{subtable}%
    \begin{subtable}{.5\linewidth}
      \centering
      \resizebox{5.8cm}{!}{
        \begin{tabular}{c|ccc|c} 
\toprule[1.5pt]
\small
\multirow{2}{*}{\textsc{NYUv2}}      & Normal & Sem Seg. & Depth & \multirow{2}{*}{$\Delta_{\text{MTL}}$$\uparrow$}\\ 
& (mDist.) $\downarrow$ &(mIoU) $\uparrow$ & (aErr.) $\downarrow$ & \\
\midrule[1.5pt]
\textsc{Single}      &22.03         & 53.81              & 38.87   & - \\ \midrule
\textsc{Vanilla} & 23.12    &54.87   & 35.09   & +6.74\%  \\
+ Ours  & 22.78     & 54.98 & 34.98 & +8.77\%  \\ \midrule[1.5pt]		
\multirow{2}{*}{\textsc{Cityscapes}}       & Sem Seg. & Part Seg. & Disparity & \multirow{2}{*}{$\Delta_{\text{MTL}}$$\uparrow$}\\ 
& (mIoU) $\uparrow$ &(mIoU) $\uparrow$ & (aErr.) $\downarrow$ & \\
\midrule[1.5pt]
\textsc{Single}       &56.08     & 53.33   & 0.74   & - \\ \midrule
\textsc{Vanilla}  & 56.02     & 54.18   & 0.69   & +2.79\%  \\
+ Ours       & 57.13     & 54.37   & 0.68   & +3.98\% \\\bottomrule[1.5pt]
\end{tabular}}
      \caption{}\label{subtab:t2}
    \end{subtable} 

    \begin{subtable}{1\linewidth}
      \centering
      \resizebox{7cm}{!}{
        \begin{tabular}{c|ccccc|c}
    \toprule[1.5pt]
\multirow{2}{*}{Method} & Seg. & H. Parts & Norm.&Sal.&Edge & \multirow{2}{*}{$\Delta_{\text{MTL}}$$\uparrow$}\\ 
& (IoU) $\uparrow$ &(IoU) $\uparrow$ & (mErr) $\downarrow$ & (IoU) $\uparrow$& (odsF)$\uparrow$ \\ \midrule[1.5pt]
    \textsc{Vanilla}    & 63.8 & 58.6 & 14.9 & 65.1&69.2&-2.86\% \\\midrule
    \textsc{+DGR}    &  64.2 & 58.7 & 14.8 & 65.2&69.5&-2.42\% \\\midrule[1.5pt]
    \textsc{MTAN}      &  63.7 & 58.9 & 14.8 & 65.4&69.6&-2.39\%    \\ \midrule
    \textsc{+DGR}   &  63.8 & 59.0 & 14.7 & 65.5&69.6&-2.19\%\\\bottomrule[1.5pt]
    \end{tabular}}
    \caption{}\label{subtab:t3}
    \end{subtable}
\end{table}

We performed additional sensitivity analyses of the proposed method under various conditions. First, we verified the performance of DGR by varying the numbers of dummy decoders and primary tasks on visual scene understanding (\textsc{NYUv2} and \textsc{Cityscapes}) as well as single-task image classification ( \textsc{IMAGENET} \cite{deng2009imagenet}). Second, we evaluated DGR with the SWIN transformer \cite{liu2021swin} backbone architecture, which is more capable than the backbone we previously used. Finally, DGR was applied to the \textsc{Pascal} dataset to verify its robustness on larger number of tasks.

\noindent \textbf{Results.}
As illustrated in Table \ref{subtab:t1}, the performance improvement of DGR compared to \textsc{VANILLA} becomes more prominent as the number of tasks increases. Moreover, we observed a slight improvement even when DGR was employed for single-task image classification (\textsc{IMAGENET}). Furthermore, Table \ref{subtab:t2} demonstrates that the SWIN backbone consistently improved performance compared to ResNet50 across all tasks in both single- and multi-task scenarios. DGR further enhanced performance, indicating its effectiveness with high capable backbone architectures. In the case of the \textsc{Pascal} dataset, shown in Table \ref{subtab:t3}, although MTL had lower performance compared to single-task learning, DGR achieved performance improvement in MTL scenarios.

These findings demonstrate that DGR performs robustly under various conditions and can achieve performance improvements when combined with more advanced backbone architectures, such as the Swin transformer.

\section{Conclusion}
We present a novel approach, Dummy Gradient norm Regularization (DGR), to improve the universality of a shared encoder in MTL. Through experiments on multiple benchmark datasets, we have demonstrated that DGR effectively improves the universality of a shared encoder, resulting in better multi-task prediction performances. Our approach takes advantage of its inherent simplicity, leading to relatively less computation time and allowing seamless integration with existing MTL algorithms. Overall, our study contributes to the advancement of MTL by addressing an important question of improving the universality of a shared encoder and providing a practical and efficient method to achieve this goal.

\section*{Acknowledgements}
This research was supported by the National Research Foundationof Korea (NRF) grant funded by the Ministry of Science and ICT (MSIT) of Korea (No. RS-2023-00208412).

\bibliographystyle{splncs04}
\bibliography{main}

\begin{thebibliography}{10}
\providecommand{\url}[1]{\texttt{#1}}
\providecommand{\urlprefix}{URL }
\providecommand{\doi}[1]{https://doi.org/#1}

\bibitem{anderson2022improving}
Anderson, C., Farrell, R.: Improving fractal pre-training. In: Proceedings of the IEEE/CVF Winter Conference on Applications of Computer Vision. pp. 1300--1309 (2022)

\bibitem{bachmann2022multimae}
Bachmann, R., Mizrahi, D., Atanov, A., Zamir, A.: Multimae: Multi-modal multi-task masked autoencoders. In: European Conference on Computer Vision. pp. 348--367. Springer (2022)

\bibitem{bengio2013representation}
Bengio, Y., Courville, A., Vincent, P.: Representation learning: A review and new perspectives. IEEE transactions on pattern analysis and machine intelligence  \textbf{35}(8),  1798--1828 (2013)

\bibitem{caruana1998multitask}
Caruana, R.: Multitask learning. Springer (1998)

\bibitem{chen2021lottery}
Chen, T., Frankle, J., Chang, S., Liu, S., Zhang, Y., Carbin, M., Wang, Z.: The lottery tickets hypothesis for supervised and self-supervised pre-training in computer vision models. In: Proceedings of the IEEE/CVF conference on computer vision and pattern recognition. pp. 16306--16316 (2021)

\bibitem{chen2018multi}
Chen, Y., Zhao, D., Lv, L., Zhang, Q.: Multi-task learning for dangerous object detection in autonomous driving. Information Sciences  \textbf{432},  559--571 (2018)

\bibitem{chen2018gradnorm}
Chen, Z., Badrinarayanan, V., Lee, C.Y., Rabinovich, A.: Gradnorm: Gradient normalization for adaptive loss balancing in deep multitask networks. In: International conference on machine learning. pp. 794--803. PMLR (2018)

\bibitem{chen2020just}
Chen, Z., Ngiam, J., Huang, Y., Luong, T., Kretzschmar, H., Chai, Y., Anguelov, D.: Just pick a sign: Optimizing deep multitask models with gradient sign dropout. Advances in Neural Information Processing Systems  \textbf{33},  2039--2050 (2020)

\bibitem{chowdhuri2019multinet}
Chowdhuri, S., Pankaj, T., Zipser, K.: Multinet: Multi-modal multi-task learning for autonomous driving. In: 2019 IEEE Winter Conference on Applications of Computer Vision (WACV). pp. 1496--1504. IEEE (2019)

\bibitem{cordts2016cityscapes}
Cordts, M., Omran, M., Ramos, S., Rehfeld, T., Enzweiler, M., Benenson, R., Franke, U., Roth, S., Schiele, B.: The cityscapes dataset for semantic urban scene understanding. In: Proceedings of the IEEE conference on computer vision and pattern recognition. pp. 3213--3223 (2016)

\bibitem{deng2009imagenet}
Deng, J., Dong, W., Socher, R., Li, L.J., Li, K., Fei-Fei, L.: Imagenet: A large-scale hierarchical image database. In: 2009 IEEE conference on computer vision and pattern recognition. pp. 248--255. Ieee (2009)

\bibitem{desideri2012multiple}
D{\'e}sid{\'e}ri, J.A.: Multiple-gradient descent algorithm (mgda) for multiobjective optimization. Comptes Rendus Mathematique  \textbf{350}(5-6),  313--318 (2012)

\bibitem{dong2021defect}
Dong, X., Taylor, C.J., Cootes, T.F.: Defect classification and detection using a multitask deep one-class cnn. IEEE Transactions on Automation Science and Engineering  \textbf{19}(3),  1719--1730 (2021)

\bibitem{du2018adapting}
Du, Y., Czarnecki, W.M., Jayakumar, S.M., Farajtabar, M., Pascanu, R., Lakshminarayanan, B.: Adapting auxiliary losses using gradient similarity. arXiv preprint arXiv:1812.02224  (2018)

\bibitem{evgeniou2004regularized}
Evgeniou, T., Pontil, M.: Regularized multi--task learning. In: Proceedings of the tenth ACM SIGKDD international conference on Knowledge discovery and data mining. pp. 109--117 (2004)

\bibitem{finn2017model}
Finn, C., Abbeel, P., Levine, S.: Model-agnostic meta-learning for fast adaptation of deep networks. In: International conference on machine learning. pp. 1126--1135. PMLR (2017)

\bibitem{foret2020sharpness}
Foret, P., Kleiner, A., Mobahi, H., Neyshabur, B.: Sharpness-aware minimization for efficiently improving generalization. arXiv preprint arXiv:2010.01412  (2020)

\bibitem{de2021part}
de~Geus, D., Meletis, P., Lu, C., Wen, X., Dubbelman, G.: Part-aware panoptic segmentation. In: Proceedings of the IEEE/CVF Conference on Computer Vision and Pattern Recognition. pp. 5485--5494 (2021)

\bibitem{guo2018dynamic}
Guo, M., Haque, A., Huang, D.A., Yeung, S., Fei-Fei, L.: Dynamic task prioritization for multitask learning. In: Proceedings of the European conference on computer vision (ECCV). pp. 270--287 (2018)

\bibitem{he2016deep}
He, K., Zhang, X., Ren, S., Sun, J.: Deep residual learning for image recognition. In: Proceedings of the IEEE conference on computer vision and pattern recognition. pp. 770--778 (2016)

\bibitem{hjelm2018learning}
Hjelm, R.D., Fedorov, A., Lavoie-Marchildon, S., Grewal, K., Bachman, P., Trischler, A., Bengio, Y.: Learning deep representations by mutual information estimation and maximization. arXiv preprint arXiv:1808.06670  (2018)

\bibitem{ishihara2021multi}
Ishihara, K., Kanervisto, A., Miura, J., Hautamaki, V.: Multi-task learning with attention for end-to-end autonomous driving. In: Proceedings of the IEEE/CVF Conference on Computer Vision and Pattern Recognition. pp. 2902--2911 (2021)

\bibitem{kendall2018multi}
Kendall, A., Gal, Y., Cipolla, R.: Multi-task learning using uncertainty to weigh losses for scene geometry and semantics. In: Proceedings of the IEEE conference on computer vision and pattern recognition. pp. 7482--7491 (2018)

\bibitem{kingma2014adam}
Kingma, D.P., Ba, J.: Adam: A method for stochastic optimization. arXiv preprint arXiv:1412.6980  (2014)

\bibitem{kolesnikov2020big}
Kolesnikov, A., Beyer, L., Zhai, X., Puigcerver, J., Yung, J., Gelly, S., Houlsby, N.: Big transfer (bit): General visual representation learning. In: Computer Vision--ECCV 2020: 16th European Conference, Glasgow, UK, August 23--28, 2020, Proceedings, Part V 16. pp. 491--507. Springer (2020)

\bibitem{kolesnikov2019revisiting}
Kolesnikov, A., Zhai, X., Beyer, L.: Revisiting self-supervised visual representation learning. In: Proceedings of the IEEE/CVF conference on computer vision and pattern recognition. pp. 1920--1929 (2019)

\bibitem{lee2022multitask}
Lee, S., Son, Y.: Multitask learning with single gradient step update for task balancing. Neurocomputing  \textbf{467},  442--453 (2022)

\bibitem{li2021end}
Li, Y., Li, J.: An end-to-end defect detection method for mobile phone light guide plate via multitask learning. IEEE Transactions on Instrumentation and Measurement  \textbf{70},  1--13 (2021)

\bibitem{lin2019pareto}
Lin, X., Zhen, H.L., Li, Z., Zhang, Q.F., Kwong, S.: Pareto multi-task learning. Advances in neural information processing systems  \textbf{32} (2019)

\bibitem{liu2021conflict}
Liu, B., Liu, X., Jin, X., Stone, P., Liu, Q.: Conflict-averse gradient descent for multi-task learning. Advances in Neural Information Processing Systems  \textbf{34},  18878--18890 (2021)

\bibitem{liu2018darts}
Liu, H., Simonyan, K., Yang, Y.: Darts: Differentiable architecture search. arXiv preprint arXiv:1806.09055  (2018)

\bibitem{liu2021towards}
Liu, L., Li, Y., Kuang, Z., Xue, J., Chen, Y., Yang, W., Liao, Q., Zhang, W.: Towards impartial multi-task learning. iclr (2021)

\bibitem{liu2019loss}
Liu, S., Liang, Y., Gitter, A.: Loss-balanced task weighting to reduce negative transfer in multi-task learning. In: Proceedings of the AAAI conference on artificial intelligence. vol.~33, pp. 9977--9978 (2019)

\bibitem{liu2019self}
Liu, S., Davison, A., Johns, E.: Self-supervised generalisation with meta auxiliary learning. Advances in Neural Information Processing Systems  \textbf{32} (2019)

\bibitem{liu2022auto}
Liu, S., James, S., Davison, A.J., Johns, E.: Auto-lambda: Disentangling dynamic task relationships. arXiv preprint arXiv:2202.03091  (2022)

\bibitem{liu2019end}
Liu, S., Johns, E., Davison, A.J.: End-to-end multi-task learning with attention. In: Proceedings of the IEEE/CVF conference on computer vision and pattern recognition. pp. 1871--1880 (2019)

\bibitem{liu2021swin}
Liu, Z., Lin, Y., Cao, Y., Hu, H., Wei, Y., Zhang, Z., Lin, S., Guo, B.: Swin transformer: Hierarchical vision transformer using shifted windows. In: Proceedings of the IEEE/CVF international conference on computer vision. pp. 10012--10022 (2021)

\bibitem{maninis2019attentive}
Maninis, K.K., Radosavovic, I., Kokkinos, I.: Attentive single-tasking of multiple tasks. In: Proceedings of the IEEE/CVF Conference on Computer Vision and Pattern Recognition. pp. 1851--1860 (2019)

\bibitem{meyerson2018pseudo}
Meyerson, E., Miikkulainen, R.: Pseudo-task augmentation: From deep multitask learning to intratask sharing—and back. In: International Conference on Machine Learning. pp. 3511--3520. PMLR (2018)

\bibitem{navon2020auxiliary}
Navon, A., Achituve, I., Maron, H., Chechik, G., Fetaya, E.: Auxiliary learning by implicit differentiation. arXiv preprint arXiv:2007.02693  (2020)

\bibitem{navon2022multi}
Navon, A., Shamsian, A., Achituve, I., Maron, H., Kawaguchi, K., Chechik, G., Fetaya, E.: Multi-task learning as a bargaining game. arXiv preprint arXiv:2202.01017  (2022)

\bibitem{rebuffi2017icarl}
Rebuffi, S.A., Kolesnikov, A., Sperl, G., Lampert, C.H.: icarl: Incremental classifier and representation learning. In: Proceedings of the IEEE conference on Computer Vision and Pattern Recognition. pp. 2001--2010 (2017)

\bibitem{ruder2017overview}
Ruder, S.: An overview of multi-task learning in deep neural networks. arXiv preprint arXiv:1706.05098  (2017)

\bibitem{sampath2023attention}
Sampath, V., Maurtua, I., Mart{\'\i}n, J.J.A., Rivera, A., Molina, J., Gutierrez, A.: Attention guided multi-task learning for surface defect identification. IEEE Transactions on Industrial Informatics  (2023)

\bibitem{sener2018multi}
Sener, O., Koltun, V.: Multi-task learning as multi-objective optimization. Advances in neural information processing systems  \textbf{31} (2018)

\bibitem{shamsian2023auxiliary}
Shamsian, A., Navon, A., Glazer, N., Kawaguchi, K., Chechik, G., Fetaya, E.: Auxiliary learning as an asymmetric bargaining game. In: International Conference on Machine Learning. pp. 30689--30705. PMLR (2023)

\bibitem{shao2022pixel}
Shao, L., Zhang, E., Ma, Q., Li, M.: Pixel-wise semisupervised fabric defect detection method combined with multitask mean teacher. IEEE Transactions on Instrumentation and Measurement  \textbf{71},  1--11 (2022)

\bibitem{silberman2012indoor}
Silberman, N., Hoiem, D., Kohli, P., Fergus, R.: Indoor segmentation and support inference from rgbd images. ECCV (5)  \textbf{7576},  746--760 (2012)

\bibitem{tzeng2017adversarial}
Tzeng, E., Hoffman, J., Saenko, K., Darrell, T.: Adversarial discriminative domain adaptation. In: Proceedings of the IEEE conference on computer vision and pattern recognition. pp. 7167--7176 (2017)

\bibitem{vandenhende2021multi}
Vandenhende, S., Georgoulis, S., Van~Gansbeke, W., Proesmans, M., Dai, D., Van~Gool, L.: Multi-task learning for dense prediction tasks: A survey. IEEE transactions on pattern analysis and machine intelligence  \textbf{44}(7),  3614--3633 (2021)

\bibitem{wang2019transferable}
Wang, X., Li, L., Ye, W., Long, M., Wang, J.: Transferable attention for domain adaptation. In: Proceedings of the AAAI Conference on Artificial Intelligence. vol.~33, pp. 5345--5352 (2019)

\bibitem{wu2018unsupervised}
Wu, Z., Xiong, Y., Yu, S.X., Lin, D.: Unsupervised feature learning via non-parametric instance discrimination. In: Proceedings of the IEEE conference on computer vision and pattern recognition. pp. 3733--3742 (2018)

\bibitem{yu2020gradient}
Yu, T., Kumar, S., Gupta, A., Levine, S., Hausman, K., Finn, C.: Gradient surgery for multi-task learning. Advances in Neural Information Processing Systems  \textbf{33},  5824--5836 (2020)

\bibitem{zhang2021understanding}
Zhang, C., Bengio, S., Hardt, M., Recht, B., Vinyals, O.: Understanding deep learning (still) requires rethinking generalization. Communications of the ACM  \textbf{64}(3),  107--115 (2021)

\bibitem{zhang2018overview}
Zhang, Y., Yang, Q.: An overview of multi-task learning. National Science Review  \textbf{5}(1),  30--43 (2018)

\bibitem{zhifei2017cvpr}
Zhang, Z., Song, Y., Qi, H.: Age progression/regression by conditional adversarial autoencoder. In: Proceedings of the IEEE conference on computer vision and pattern recognition. pp. 5810--5818 (2017)

\bibitem{zhang2022task}
Zhang, Z., Wang, S., Xu, Y., Fang, Y., Yu, W., Liu, Y., Zhao, H., Zhu, C., Zeng, M.: Task compass: Scaling multi-task pre-training with task prefix. arXiv preprint arXiv:2210.06277  (2022)

\end{thebibliography}
\end{document}